\documentclass[conference]{IEEEtran}
\IEEEoverridecommandlockouts
\usepackage{cite}
\usepackage{amsmath,amssymb,amsfonts}
\usepackage{algpseudocode}
\usepackage{graphicx}
\usepackage{textcomp}
\usepackage{xcolor}
\def\BibTeX{{\rm B\kern-.05em{\sc i\kern-.025em b}\kern-.08em
    T\kern-.1667em\lower.7ex\hbox{E}\kern-.125emX}}
\usepackage{amssymb}
\usepackage{amsmath}
\usepackage[ruled,linesnumbered]{algorithm2e}
\usepackage{multirow}
\usepackage{amsmath}
\usepackage{graphics}
\usepackage{makecell}
\usepackage{epsfig}
\usepackage{etoolbox}
\makeatletter
\patchcmd{\@makecaption}
  {\scshape}
  {}
  {}
  {}
\makeatother
\begin{document}

\title{Trainable Class Prototypes for Few-Shot Learning \\
}

\author{\IEEEauthorblockN{Jianyi Li}
\IEEEauthorblockA{\textit{School of Information and Communications} \\
\textit{Xi’an Jiaotong University}\\
Xi'an, P.R. China \\
lijianyi1488@stu.xjtu.edu.cn}
\and
\IEEEauthorblockN{Guizhong Liu}
\IEEEauthorblockA{\textit{School of Information and Communications} \\
\textit{Xi’an Jiaotong University}\\
Xi'an, P.R. China \\
liugz@xjtu.edu.cn }
 }

\maketitle

\begin{abstract}
Metric learning is a widely used method for few shot learning in which the quality of prototypes plays a key role in the algorithm. In this paper we propose the trainable prototypes for distance measure instead of the artificial ones within the meta-training and task-training framework. Also to avoid the disadvantages that the episodic meta-training brought, we adopt non-episodic meta-training based on self-supervised learning. Overall we solve the few-shot tasks in two phases: meta-training a transferable feature extractor via self-supervised learning and training the prototypes for metric classification. In addition, the simple attention mechanism is used in both meta-training and task-training. Our method achieves state-of-the-art performance in a variety of established few-shot tasks on the standard few-shot visual classiﬁcation dataset, with about 20\% increase compared to the available unsupervised few-shot learning methods.
\end{abstract}

\section{Introduction}
In recent years deep learning has made major advances in computer vision areas such as image recognition, video object detection and tracking. A deep neural network needs a large amount of labeled data to fit its parameters whereas it is laborious to label so many examples by human annotators. Thus the problem of learning with few labeled samples called few-shot learning has been paid more and more attention. Few-shot learning is described as a classification task set in N -way and k -shot, which means to distinguish N categories, each of which has k (quite small) labeled samples. The model predict classes for new examples only depending on k labeled data. The annotated data is called the support set, and the new data belonging to the N categories is called query set.

\begin{figure}[htbp]
\centerline{\includegraphics[width=10cm, height=4.55cm]{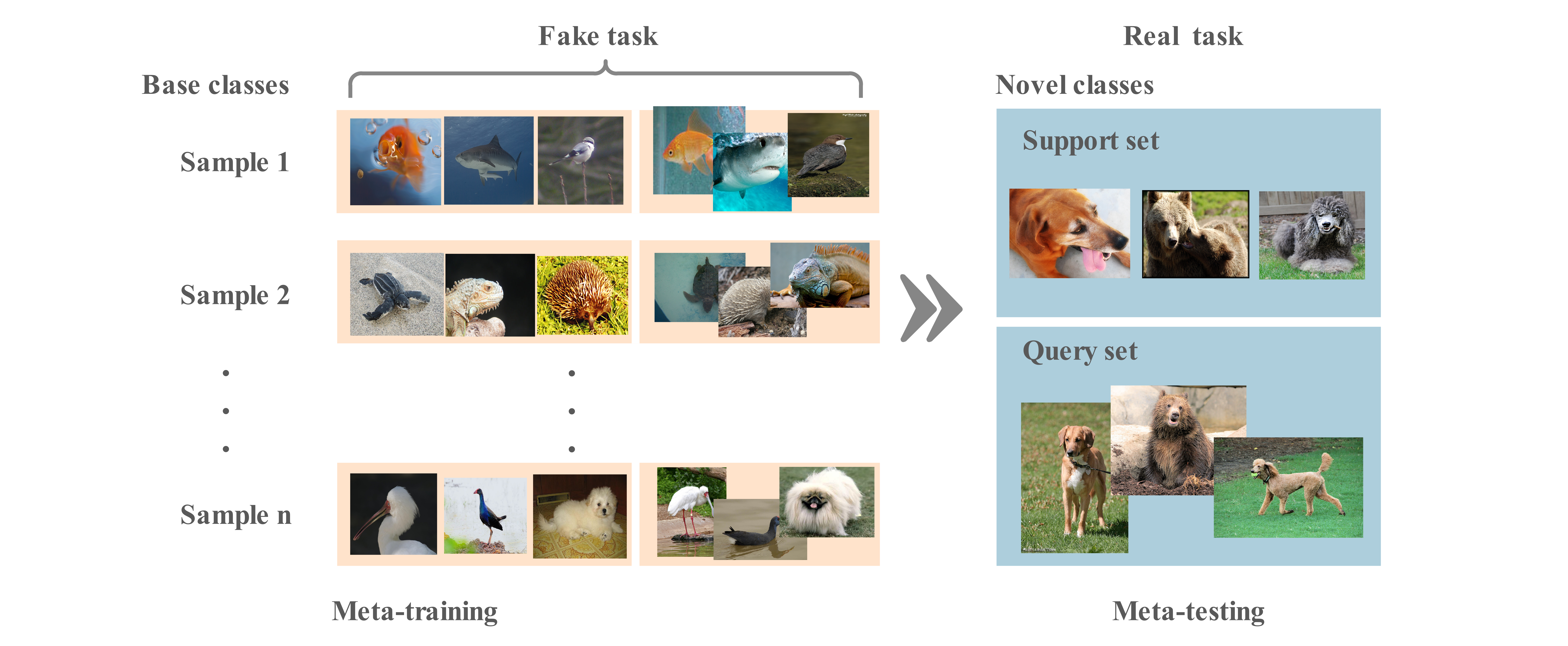}}
\caption{The universal method used in supervised few-shot learning, which consists of meta-training and meta-testing. In the meta-training, the training sample is actually a mimic few-shot task comprised of some labeled data chosen from base classes. And in the meta-testing the model will solve a real task with few labeled data and an unlabeled query set chosen from novel classes. We show a model trained for solving 3-way 1-shot tasks in this figure.}
\label{fig1}
\end{figure}

People have proposed varieties of few-shot methods, all of which rely on meta-training assisted with base classes. The universal approach is to use the base classes to construct fake few-shot tasks for training the network first, with the purpose of enabling the network an ability to accomplish real few-shot tasks through simulating the process of carrying out the fake tasks. This is called the meta-training stage with tasks as samples. Next, use the trained network to complete real few-shot tasks of novel classes, and calculate the classification accuracy on the query set in the tasks to evaluate the algorithm, which is usually called the meta-testing. The whole procedure is shown in Fig.\ref{fig1}.

Few-shot learning algorithms could be classified into three categories. The first \cite{vinyals2016matching,snell2017prototypical,sung2018learning,oreshkin2018tadam} is based on metric learning, which consists of three steps of feature extraction, distance measure, and prediction, relying on effective metric design and reducing the cross entropy loss in meta training to improve classification accuracy. The second are the teacher-student network based methods including \cite{finn2017model,rusu2018meta,li2019lgm,ravi2016optimization}. The teacher network guides the student network to solve the few-shot tasks in terms of parameter initialization, parameter update and other aspects. The algorithm enables the teacher network to obtain the ability to instruct the student network via meta-training. The third category such as \cite{liu2018learning} and \cite{kim2019edge} is based on the transduction, which propagates the label of the support data to the queries through a specific graph, thereby obtaining the predicted class of the query set. The algorithm optimizes the accuracy of propagation label through meta-training.

The meta-training determines the model’s performance in the few-shot learning algorithm. However it brings two obvious drawbacks. First, the meta-training requires a large number of labeled auxiliary examples (base classes). Those algorithms can not work without adequate labeled samples. Second, the meta-training phase uses tasks as training samples. Therefore, a task type decided by the values of N and k needs to be certain before meta-training to ensure that the number of images contained by each mimic few shot task (i.e. a meta-training sample) is consistent during training. The meta-trained network can only be used to solve few-shot tasks with the same type as the meta-training samples, and it performs worse in other types of tasks. However, in reality we need to solve various types of few-shot tasks, and it is unreasonable to meta-train the network from scratch in order to solve a few shot task with new type.

In order to solve these two problems, we solve the few shot classification based on self-supervised learning according to \cite{li2020few}. Specifically, our method called  TrainPro-FSL abandons the episodic meta-training phase, which takes the fake few-shot tasks as samples, and uses instead two new phases: the non-episodic meta-training via self-supervised learning directly using a single image as a training sample, and the task-training to solve a real few shot task. In the first phase, a discriminative self-supervised learning method is used to obtain a feature extractor with good generalization ability using unlabeled images. In the second phase, we use metric learning to solve real few-shot tasks. The meta-trained feature extractor is used to extract features from all the images in the current task, and SGC is carried out, based on a specific graph defined by the current task as the feature aggregation method described in \cite{li2020few}. We use the aggregated support set features to obtain the class prototypes which are trained via a simple neural network. Finally we use the class prototypes to implement distance measure and further classify the query images.

Furthermore, we demonstrate that the attention mechanism are helpful for both meta-training and task-training. We add attention structure to the neural network in the meta-training phase. In the task-training phase, an attention mask that does not increase network parameters is introduced to correct the query image features. Both of these methods can improve the classification accuracy of our algorithm.

Our key contributions can be summarized as follows:
\begin{itemize}
\item We propose TrainPro-FSL, a  unsupervised few-shot learning algorithm with trainable prototypes. By adopting the methodology of self-supervised learning, the two problems intrinsic in the existing episodic meta-training paradigm are solved simultaneously. Thus our method does not require a large number of labeled samples for training. In addition, the meta-trained model can carry out different types of real few-shot tasks.
\item We propose to use trainable prototypes to implement metric classification, which is better than the manual design in \cite{wu2019simplifying}.
\item We propose to use a simple attentive mechanism in the metric learning to further correct the feature of examples in the query set.
\item Adequate experiments demonstrate that our method reaches state-of-the-art accuracy on miniImageNet , a standardized benchmark in few-shot learning.
\end{itemize}

The paper is organized as follows. In \ref{Section 2}, we introduce the related works. Our methodology is described in \ref{Section 3}. In \ref{Section 4}, experimental results on the standard vision dataset are shown in comparison with the proposed works. Finally, a conclusion is drawn in \ref{Section 5}.

\section{Related Works}\label{Section 2}
In this section we aim to show the metric learning used in few-shot learning algorithms proposed in previous years. In addition, we introduce some unsupervised few-shot learning methods presented recently and popular self-supervised algorithms.

\subsection{Metric Learning}
The core of metric learning is to extract features from the support set and query set, then obtain the class prototypes using the support set, and predict classes of queries via the nearest neighbor algorithm and attention mechanism. Therefore the quality of class prototypes obtained in algorithm are vital for the performance of the metric learning. Through episodic meta-training, a metric based method obtains a feature extractor that facilitates completing the classification task based on distance measurement. Matching Networks \cite{vinyals2016matching} used LSTM to extract full context embeddings from images and applied attention mechanism to classify. Prototypical Networks \cite{snell2017prototypical} proposed to use Euclidean distance to better measure the similarity between features, and use prototypes of each class to classify queries. Relation Network \cite{sung2018learning} used a neural network to replace the traditional distance metric, and directly output the queries’ categories via an end-to-end network. DC-IMP \cite{lifchitz2019dense} introduced dense classification and leverage implanting to bring metric learning the task dependency.

\subsection{Unsupervised Few Shot Learning}
The base classes in unsupervised methods has no labels. Some existing methods use unsupervised learning or data enhancement methods to leverage these unlabeled base classes to artificially construct fake support set and query set for meta training. They are able to combine with the few-shot learning methods as mentioned above (such as MAML \cite{finn2017model} and Prototypical Net \cite{snell2017prototypical}) to fulfil few-shot tasks. UFLST \cite{ji2019unsupervised} and CACTU \cite{hsu2018unsupervised} use clustering to make pseudo-labels for unlabeled examples, then use the pseudo-labeled data as ordinary labeled data to construct fake few-shot tasks to complete meta-training.  AAL \cite{antoniou2019assume} and UMTRA \cite{khodadadeh2019unsupervised} took each instance as one class and randomly sample multiple examples to construct a fake support set, then generate a corresponding query set according to the support set by data augmentation techniques. ULDA \cite{qin2020unsupervised} developed a new simple data augmentation method to enhance the difference between the support set distribution and query set distribution when constructing the fake few-shot tasks for meta-training.

\subsection{Self-Supervised Learning}
The popular self-supervised algorithms can be divided into three categories, namely contrastive self-supervised learning, generative self-supervised learning and discriminative self-supervised learning. The core of contrastive self-supervised learning is contrastive learning. The main idea of contrastive learning is to make the features of positive sample pairs close to each other in the feature space, and keep the features of negative sample pairs away from each other, such as BYOL\cite{grill2020bootstrap}, SimCLR\cite{chen2020simple}, and their performance can even be comparable to supervised algorithms. Generative self-supervised learning algorithms such as image colorization\cite{zhang2016colorful} and VQ-VAE\cite{oord2017neural}, usually take the original image as the groundtruth, train the network and make the network output close to the groundtruth. In generative self-supervised learning, images are generated at the pixel level, so the features extracted by the network are partial to image details, rather than semantic features. 

Discriminant self-supervised learning can train the classification algorithm without the category label. The category here is set by the algorithm itself, and the label of the training sample is automatically generated according to the algorithm setting. In the discriminative self-supervised algorithm, the design of pretext tasks is diverse. Commonly used algorithms include jigsaw puzzle\cite{noroozi2016unsupervised}, 
rotation prediction\cite{gidaris2018unsupervised}, and deep clustering \cite{tian2017deepcluster}.

\section{Methodology}\label{Section 3}
The notations and problem formulation of self-supervised few-shot learning are introduced in \ref{3.1}, and our paradigm is presented in \ref{3.2}. Finally, the loss design for trainable prototypes and attention mechanism are described in \ref{3.3} and \ref{3.4} respectively. 
\subsection{Problem Formulation}\label{3.1}
Given two datasets, namely $D_{base}$ and $D_{task}$ with disjoint classes. $D_{base}$ consists of a large number of unlabeled examples from the base classes. $D_{task}$ has a small number of labeled examples called the support set $D_s$, along with some unlabeled ones called the query set $D_q$, all from the new classes. They stand for the total data in a few-shot learning task. The number of classes in the novel dataset $D_{task}$, the number of support samples and the number of query inputs for each of these classes are denoted $N$, $k$ and $q$ respectively. So there are totally $N\times(k+q)$ examples in a few-shot learning task. Our aim is to predict the classes of the query set of $D_{task}$. 
Different from the previous works like \cite{ravi2016optimization}, our $D_{base}$ does not have any labels. So we train the classification network with only a few labeled examples namely $D_s$ in a real sense.

\subsection{Proposed Paradigm of Solution}\label{3.2}
We first train a backbone deep neural network able to extract useful and compact features from inputs, which will be used as a generic feature extractor. In this non-episodic meta-training phase, we train the network with $D_{base} = \left\{ x_{1}^{'},x_{2}^{'},......,x_{n}^{'} \right\}~$  where $x^{'} \in R^{w \times h \times 3}$ via ODC in \cite{tian2019contrastive}, a kind of discrimitive self-supervised learning method,  which promises a transferable feature extractor. Thus we obtain the extractor $\left. f_{\varphi}:R^{w \times h \times 3}\rightarrow R^{e} \right.$.

We then use $f_\varphi$ to obtain the features of the total data in $D_{task}$ (both $D_s$ and $D_q$) namely $f_{\varphi}\left( D_{task} \right) = \left\{ f_{\varphi}\left( x \right) \middle| x \in D_{task} \right\}$. Then we step into the second phase namely the task-training phase. First we build a nearest neighbor graph using the cosine similarity according to \cite{hu2020exploiting} :
\begin{equation}
cos\left( {f_{\varphi}\left( x_{1} \right),f_{\varphi}\left( x_{2} \right)} \right) = \frac{f_{\varphi}\left( x_{1} \right)^{T}f_{\varphi}\left( x_{2} \right)}{\left\| {f_{\varphi}\left\| x_{1} \right\|} \right\|_{2}\left\| {f_{\varphi}\left\| x_{2} \right\|} \right\|_{2}}
\end{equation}
The base graph denoted $G_{task}\left( V,E \right)$ uses $f_\varphi(D_{task})$ to construct vertices. In details, its vertices matrix $V \in R^{\lbrack N \times (k + q)\rbrack \times e}$ is the stacked representations of support set and query set i.e. each vertex represents an image's feature. We make the values of graph edges represent the similarity between vertices-that is, similar vertices have larger adjacency values. To get the adjacency matrix $E \in R^{\lbrack N \times (k + q)\rbrack \times \lbrack N \times (k + q)\rbrack}$, we first define a similarity matrix $S$ with the same dimension computed as follows:
\begin{equation}
S_{i,j} = \begin{cases}
\left. {\mathit{\cos}(}V_{i,:},V_{j,:} \right) & {i \neq j} \\
0 & {i = j} \\
\end{cases}
\end{equation}
where $V_{i,:}$ denotes the $i$ -th row in $V$. Then we just save the $m$ largest values on each row and on the corresponding column in S to obtain a more sparse matrix helpful to reduce the interference. Finally, we normalize the resulting matrix to get the adjacency matrix:
\begin{equation}
E = D^{- \frac{1}{2}}SD^{- \frac{1}{2}}
\end{equation}
where $D$ is the degree diagonal matrix computed by $D_{i,i}=\sum_{j} S_{i,j}$. We can consider $E$ as the Laplacian matrix in GCN\cite{kipf2016semi} used to aggregate information among vertices.

We aggregate features for each vertex via the graph structure to get $V^{new}$ according to \cite{li2020few}. Then we use the metric learning to classify the query images based on $V^{new}$. In metric learning, the quality of class prototypes has a great influence on classification accuracy. Therefore, we abandon the traditional method of artificially constructing class prototypes and use neural networks to obtain good prototypes via training to implement the metric classification. In order to improve the performance, we adopt the attention mechanism in training phase in the meanwhile.

\begin{figure*}[htbp]
\centering
\includegraphics[width=\linewidth,scale=1.00]{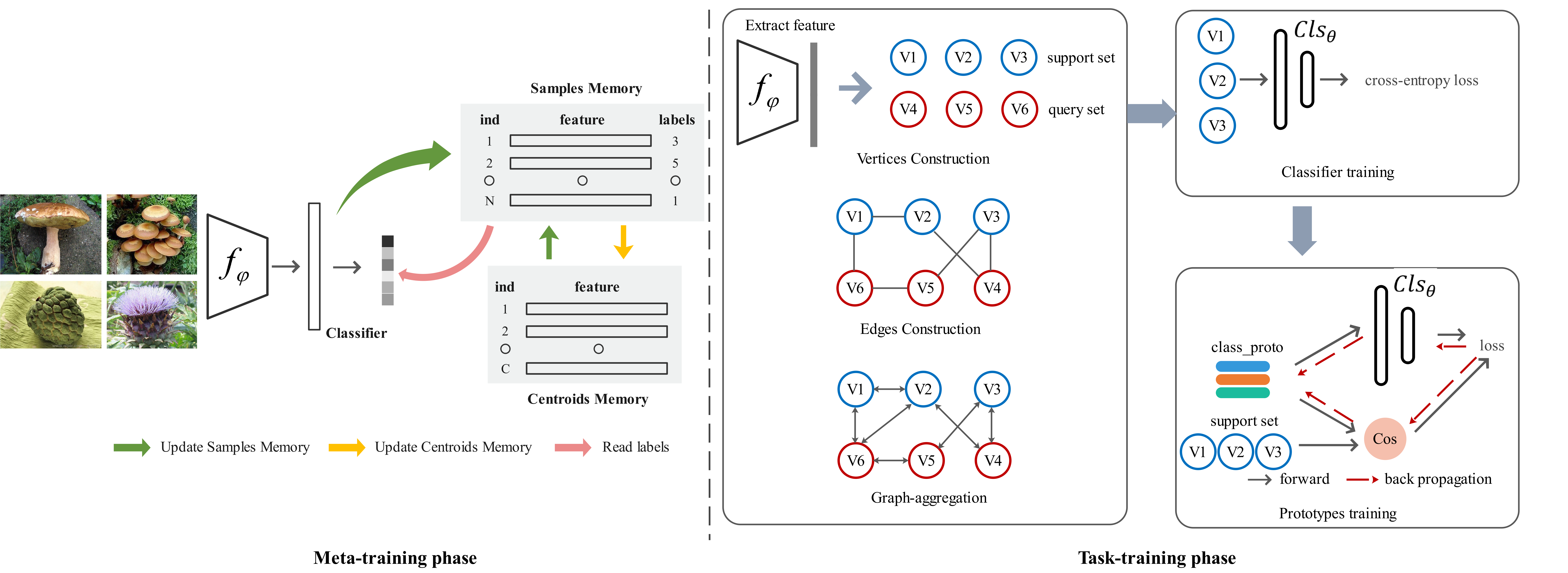}
\caption{The overall architecture of the proposed paradigm. The left shows the meta-training phase using ODC resulting in a global feature extractor. The right is the task-training phase comprised of graph-aggregation, classifier training with the support set and the prototypes training.}
\label{fig2}
\end{figure*}


\begin{algorithm}[ht]
\caption{The process of task-training phase with prototypes training.}
\label{algorithm1}
\KwIn{A $N$ -way $k$ -shot task with the dataset $D_{task}=\left\{ D_{s},D_{q} \right\}$; The meta-trained feature extractor $f_\varphi$}
\KwOut{Trained prototypes}
Obtain features of all inputs including labeled and unlabeled ones, $f_{\varphi}\left( D_{task} \right)$\;
Build the graph $G_{task}(V,E)$ based on $f_\varphi(D_{task})$\;
Aggregate vertices features $V$ of the graph to get $V^{new}$\;
randomly initialize $\theta$; \qquad\qquad\qquad\qquad\qquad\qquad\textbf{Classifier training}\;
Use manifold augmentation to extend labeled data in semantic level according to \cite{li2020few}, and get the augmented feature set $V_{aug}^{new}$\;
Train $Cls_\theta$ using $V_{aug}^{new}$ and cross entropy loss to obtain $Cls_{\theta^*}$;\qquad\qquad\qquad\qquad\qquad\qquad\qquad\qquad\textbf{Prototypes training}\;
Initialization the prototypes $proto^0$ randomly\;
Use $Cls_{\theta^*}$ to calculate the loss of the prototypes classification and the metric classification of the support set according to Eq. (\ref{eq1})\;
Update $proto^0$ based on the loss function and finally obtain $proto^*$\;
\textbf{Return} $proto^*$;

\end{algorithm}


Our paradigm is illustrated in Fig.\ref{fig2}. In general TrainPro-FSL has two phases: (1) Meta-training phase: training a generic feature extractor via ODC algorithm. (2) Task-training phase: training the prototypes  using the support set data after the feature aggregation through graph. Once the latter is finished, the performance of this model is evaluated by measuring the distance between features of $D_q$ and prototypes. The process of task-training phase is formalized in Algorithm \ref{algorithm1}.

The details of the two phases are provided in the following, ﬁrst the meta-training phase then the task-training phase.

\textbf{Meta-training phase:} We follow the methodology called ODC, an effective discrimitive self-supervised learning method, proposed in \cite{zhan2020online}. The ODC is an online update deep clustering algorithm, which improves the DeepCluster \cite{tian2017deepcluster} in the way of pseudo-label update. It discards the clustering after each epoch of training, and chooses to update the image features and pseudo-labels at the same time during the training process. ODC adds sample memory and centroids memory to DeepCluster to achieve the online update of pseudo-labels.

In a training process of ODC, there are three main steps: update network parameters, update sample memory and update centroids memory. During training we adopt the simple cross entropy loss as the loss function. Through the self-supervised learning we obtain a feature embedding, which can get features with semantic information.

\textbf{Task-training phase:} In this phase we aim to obtain the well trained prototypes for the distance measure. Therefore we need to train a classifier firstly as the base neural network to train the prototypes. Specifically we ﬁx the meta-trained parameters in the backbone and train a task-dependent classiﬁer $Cls_\theta$ on the transferred representations of the few-shot task’s dataset namely $D_{task}$. Before training a linear classiﬁer with $D_s$ having few labeled examples, a method similar to simpliﬁed graph convolution \cite{wu2019simplifying}, namely the graph aggregation is used according to \cite{li2020few}. We construct the graph $G_{task}(V,E)$ for the current few-shot task through the steps introduced before.

Then we propagate feature (\cite{wu2019simplifying}) to obtain new features for each vertex:
\begin{equation}
V^{new} = \left( \alpha I + E)^{\gamma}V \right.
\end{equation}
where $I$ is the identity matrix and $\gamma$ is a hyperparameter  denoting the number of times to aggregate feature. At the same time, $\alpha$ is also a key value to balance between the neighbors representations and the self-ones.

After aggregation, we use the labeled part of the vertices to train the task-dependent classifier $Cls_\theta$, a simple fully connected network. Then we use the trained network  $Cls_\theta$ with loss function specially designed for prototypes training to do the backpropagation and further obtain the trained prototypes. The design for loss function is introduced in details in \ref{3.3}. After finishing the task-training, we have the helpful class prototypes to do distance measure. As done in the method of usual metric learning algorithms, we classify the query image into the class corresponding to the prototype with the closest distance to the feature of this query example. To improve the classification accuracy we design a simple attention mask add to the query examples' features before the calculation of distance which is described in \ref{3.4}.
       
\subsection{Loss Design for Trainable Prototypes}\label{3.3}
In the metric learning algorithm, the quality of the prototypes determine the classification accuracy of the distance metric, but the method of directly averaging the support examples features as in the \cite{snell2017prototypical} cannot achieve the best results. We proposes trainable prototypes and consider it as a network parameter to participate in the training, so that you can get the prototypes that cannot design artificially. During the prototypes training, the design of loss function is very important. There are two requirements for the good prototypes: one is that it can be correctly classified by a trained classification head, and the other is that it can effectively improve the accuracy of metric classification. According to these two aspects, the corresponding loss function can be designed respectively.

The first term of the total loss is the cross-entropy loss of prototypes classification. In the calculation loss, the groundtruth of the class prototype is the one-hot form of its corresponding category which is a vector of length N, each category corresponds to a dimension in the vector, the value is 1 in the dimension representing the category to which it belongs, and the value is 0 in other dimensions. The classification loss can be described as follows:
\begin{equation}
L_{class} = \frac{1}{N}{\sum\limits_{i = 1}^{N}{- \frac{1}{N}\left( {\sum\limits_{c = 1}^{N}{y_{i,c}log\left( {\hat{y}}_{i,c} \right)}} \right)}}
\end{equation}
where ${\hat{y}}_{i,c}$ is the value of the prediction vector of the $i$-th class prototype in the $c$-th dimension, $y_{i,c}$ is the value on the $c$-th dimension of the one-hot label of the $i$-th class prototype. 

In order to make the training process converge, it is not enough to use only the cross entropy loss, and an entropy loss of the prototypes classification result must be designed. Because the last layer of the classification head is the softmax layer, so that the classification result is a probability distribution. The value of each dimension represents the probability that the prototypes belongs to the corresponding class. We believe that the higher credibility of the classification result of the prototype is better. Therefore, the value of the classification result in the corresponding dimension of the correct category should be close to 1, and the value of other positions should be close to 0. In order to strengthen this trend, Entropy loss can be designed as follows:
\begin{equation}
L_{entropy} = \frac{1}{N}{\sum\limits_{i = 1}^{N}{- \frac{1}{N}\left( {\sum\limits_{c = 1}^{N}{{\hat{y}}_{i,c}log\left( {\hat{y}}_{i,c} \right)}} \right)}}
\end{equation}

Entropy loss hopes that the smaller the entropy of the prediction result, the better, which ensures that the prediction result tends to the one-hot vector and strengthens the training tendency. In addition to the above two losses, cross-entropy loss needs to be calculated for the metric classification results of the support examples, which is called metric loss. This loss is to ensure that the class prototypes have a good distance metric classification ability. Since the query examples has no label, the support examples are used instead in calculating the metric loss during training. We write the metric loss as:

\begin{equation}
\begin{split}
L_{metric} = \frac{1}{N \times S}{\sum\limits_{i = 1}^{N \times S}}
\\{{- \frac{1}{N}\left( {\sum\limits_{c = 1}^{N}{y_{i,c}log\left( softmax\left( {\cos\left( {V_{i,:}^{new},~{proto}_{c}} \right)} \right) \right)}} \right)}}
\end{split}
\end{equation}
where the ${proto}_{c}$ represents the prototype of the $c$-th category. The total loss is the weighted summation of these three items expressed as follows:
\begin{equation}\label{eq1}
 L_{total} = \lambda L_{entropy} + \delta L_{class} + L_{metric} 
\end{equation}

\subsection{Attention Mechanism}\label{3.4}
We aim to pay more attention to important features while ignoring other content that are not vital for distance measure by adding attention mechanism to metric learning. Therefore, we introduce the attention mask used to correct the query examples features. Because the number of labeled support examples at task-training phase is small, the network parameters involved in this process cannot be increased, otherwise it will cause serious overfitting. The purpose of adding an attention mechanism in the metric classification is to highlight the features channel related to a certain category in both the query example feature and the corresponding prototype, so as to obtain the corrected query feature that is easier to classify. For this purpose, it is found that the class prototype trained in task-training is a natural attention mask.

The channel corresponding to a larger value in a prototype represents an important feature that distinguishes this class from other classes. Multiplying the prototype as the attention mask with the query example feature can highlight the channel related to this class in the query feature. If the query image belongs to this category, multiply the mask and then do the distance measure to get a larger similarity value, which can be better distinguished from other categories; if the query image does not belong to this category, then the similarity value will be smaller, and a better classification accuracy can also be achieved.

Therefore the attention mask are proposed as follows:
\begin{equation}
{atten\_ mask}_{n} = Softmax\left( \mu * abs\left( {proto}_{n} \right) \right)
\end{equation}
where ${atten\_ mask}_{n}$ represents the attention mask corresponding to the $n$-th category and $\mu$ is a scale factor. Then we use the attention mask to correct the query example features:
\begin{equation}
{query}_{n} = \varepsilon * query*{atten\_ mask}_{n} + query
\end{equation}
where ${query}_{n}$ is the query example features after correcting by the attention mask of the $n$-th category. $query$ is the aggregated query features. $\varepsilon$ is a weighting factor set as 10000 in our experiments.

In addition to adding the attention mask in the task-training phase, we also add the attention module SENet\cite{hu2018squeeze} to the network in the meta-training phase, and verify its effectiveness through experiments.

\section{Experiments}\label{Section 4}

We conduct experiments on the widely used few-shot image classification benchmark: miniImageNet \cite{vinyals2016matching}, which is a derivative of ImageNet. 
\subsection{Models and implementation details}
\textbf{Architecture.} In the meta-training phase we use ResNet50\cite{he2016deep} as the structure of feature extractors $f_{\varphi}$. This backbone has 50 convolutional layers grouped into 16 blocks. And in every block we add a SE block. We set the input size as $224\times224$ and flatten the outputs as inputs to the graph aggregation, so that $e=2048$ in $V \in R^{\lbrack N \times (k + q)\rbrack \times e}$.

In consideration of the extreme few labeled examples we take only one fully connected layer and a following softmax layer as the structure of the classifier $Cls$ to avoid overfitting. And in the metric classification we measure the feature distance by calculating the cosine between two vectors.

\textbf{Optimization and hyper-parameters setup.} For the meta-training phase, we train the backbone in a total of 400 epochs from scratch using the SGD optimizer \cite{bottou2010large} and the cross-entropy loss. In order to get a better model we also adopt early stopping. We set the total clusters number as 1680 while the unlabled samples used in ODC composed of 168 categories. For the task-training phase, we train the classifier and the prototypes in 11 epochs 1000 epochs respectively, using the Adam optimizer \cite{kingma2014adam} and the loss function shown in the previous section. 

\begin{table}[htbp]
\caption{Performance of TrainProto-FSL in comparison to the previous works on miniImageNet on 5-way 1-shot and 5-way 5-shot tasks. Average accuracies are reported with 95\% conﬁdence intervals.}
\begin{center}
\begin{tabular}{p{1.2cm}p{3.4cm}p{1.4cm}<{\centering}p{1.1cm}<{\centering}}
\hline
\multicolumn{2}{c}{}&\multicolumn{2}{p{2.5cm}<{\centering}}{\textbf{5-way Accuracy}} \\
\hline
\multicolumn{2}{c}{\textbf{miniImageNet}}& \textbf{1-shot} &\textbf{5-shot} \\
\hline
&\textbf{CACTUs-MAML \cite{hsu2018unsupervised}}	&39.90$\pm$0.74\%	&53.97$\pm$0.70\% \\
&\textbf{CACTUs-ProtoNets \cite{hsu2018unsupervised}}&	39.18$\pm$0.71\%	&53.36$\pm$0.70\%\\
&\textbf{UFLST \cite{ji2019unsupervised}} &33.77$\pm$0.70\%	&45.03$\pm$0.73\%\\
&\textbf{UMTRA \cite{khodadadeh2019unsupervised}} &39.93$\pm-$\%	&50.73${\pm-}$\% \\
\multirow{2}{*}{\textbf{unsupervised}} & \textbf{AAL-ProtoNets \cite{antoniou2019assume}}	&37.67$\pm$0.39\%	&40.29$\pm$0.68\%\\
&\textbf{AAL-MAML++ \cite{antoniou2019assume}}	&34.57$\pm$0.74\%	&49.18$\pm$0.47\%\\
&\textbf{ULDA-ProtoNets \cite{qin2020unsupervised}}	&40.63$\pm$0.61\%	&55.41$\pm$0.57\%\\
&\textbf{ULDA-MetaOptNet \cite{qin2020unsupervised}}	&40.71$\pm$0.62\%	&54.49$\pm$0.58\%\\
&\textbf{CSSL-FSL\_Image168\cite{li2020few}}	
&54.17$\pm$1.31\%	&68.91$\pm$0.90\%\\
&\textbf{TrainProto-FSL(ours)}	&\textbf{58.92$\pm$0.91\%}	&\textbf{73.94$\pm$0.63\%}\\

\hline
\multirow{2}{*}{\textbf{supervised}} &\textbf{MAML}	&46.60$\pm$0.74\%	&60.00$\pm$0.71\%\\
&\textbf{ProtoNets}	&47.01$\pm$0.72\%	&67.90$\pm$0.76\%\\
\hline
\end{tabular}
\label{tab1}
\end{center}
\end{table}

\subsection{Results on miniImageNet}
The miniImageNet dataset consists of 100 classes randomly sampled from the ImageNet and each class contains 600 images of size $84\times84$. It is usually divided into three parts \cite{ravi2016optimization}: training set with 64 base classes, validation set with 16 classes, and testing set with 20 novel classes. In the meta-training phase we use 168 classes randomly chosen from ImageNet according to \cite{li2020few} to get a well feature extractor, having no labels. In the task-training phase we sample novel classes to design few-shot tasks as inputs. We ensure that the novel classes have never been seen in the meta-training phase.

We evaluate our method on 1000 randomly sampled tasks and report their mean accuracy in TABLE \ref{tab1}. We compare our method in both 5-way 1-shot and 5-way 5-shot setting with some classical supervised few-shot learning methods and novel unsupervised methods proposed recently. It can be found that our method is much better than previous unsupervised few-shot learning methods( \cite{ji2019unsupervised} etc.), improving them by more than 18\%.
The CSSL-FSL\cite{li2020few} also use non-episodic meta-training to train a feature extractor via a contrastive self-supervised algorithm. Our method improves it by more than 4\%. Even compared with supervised methods(\cite{finn2017model} and \cite{snell2017prototypical}), our method still has improvement by 2-11\% both on 5-way 1-shot and 5-way 5-shot tasks.

\subsection{Ablation experiments}
In this section, we conduct ablation experiments to analyze how trainable prototypes and attention mechanism affects the few-shot image classification performance. 
TABLE \ref{tab2} shows the results of the ablation studies on miniImageNet in 5-way 1-shot, 5-way 5-shot, 5-way 10-shot and 5-way 30-shot setting. In the table, without TrainProto means that the mean value of support features is directly calculated as the prototype to participate in the distance measure and without AttenMask means don't use attention mask to correct the query features. From the table we can see the original model performs best. Attention mask before the distance measure improves accuracy by 0.1\%. When the number of shots is 1 or 5, the improvement in accuracy brought by the AttenMask is more obvious, indicating that the attention mask can achieve greater advantages when there are few label samples.

The SENet used in feature extractor can provide about 1\% extra gain. And without TrainProto, the result will decrease by 2-6\%. The reason is that TrainProto is actually equivalent to obtaining the class prototypes through a very complicated function, which cannot be designed manually. Therefore, the prototypes obtained by training must be better than those obtained by directly averaging.

\begin{table*}[htbp]
\caption{Results of ablation studies on miniImageNet. The meta-training dataset consists of 168 classes from ImageNet for all the four models.}
\begin{center}
\begin{tabular}{c c c | c c c c}
\hline
\textbf{SENet}&\textbf{AttenMask}&\textbf{TrainProto}&\textbf{1-shot}&\textbf{5-shot}&\textbf{10-shot}&\textbf{30-shot} \\
\hline
$ $&$ $&$\surd$&57.90$\pm$0.86\% &72.55$\pm$0.62\% &77.22$\pm$0.52\%&82.96$\pm$0.41\% \\
$\surd$& &$\surd$ &58.82$\pm$0.91\% &73.83$\pm$0.63\% &77.13$\pm$0.55\%&83.09$\pm$0.43\%\\
$\surd$&$\surd$& &\textbf{59.00$\pm$0.70\%} &71.77$\pm$0.52\% &74,54$\pm$0.47\%&77.02$\pm$0.43\% \\
$\surd$&$\surd$ &$\surd$&58.92$\pm$0.91\% &\textbf{73.94$\pm$0.63\%} &\textbf{77.22$\pm$0.55\%}&\textbf{83.10$\pm$0.43\%} \\
\hline
\end{tabular}
\label{tab2}
\end{center}
\end{table*}

\subsection{Other Experiments}
In this section, we compare the effects of two different attention modules on meta-training. We added Non-Local(NL) \cite{wang2018non} and SENet modules(SE) to the ODC backbone network respectively, and compared their effects on the accuracy of few shot classification after training. The results are shown in TABLE \ref{tab3}. We also compares the algorithm without attention module(w/o Atten) in meta-training with the two improved algorithms mentioned above.

Generally speaking, the addition of SENet and Non-local modules can improve the performance of the algorithm, and the improvement of SENet is more than that of Non-local. On the few-sample classification tasks with shot=1 and 5, the two attention modules brought the higher improvement in classification accuracy, reaching a maximum of 1.2\%. But when shot=10, the SENet module brings a 0.06\% decrease in accuracy, and when shot=30, the Non-local module brings a 0.38\% decrease in accuracy. In general, these two attentions have played a positive role in the training of the feature extractor, and the improvement of SEnet is more obvious especially when shot is extreme small.

\begin{table}[htbp]
\caption{Performance of TrainProto-FSL with different attention modules on miniImageNet on different types of tasks. For each type of task, the best-performing method is in bold.}
\begin{center}
\begin{tabular}{l c c c c c}
\hline
\textbf{5-way Accuracy}&\textbf{1shot}&\textbf{5shot}&\textbf{10shot} &\textbf{20shot} &\textbf{30shot} \\
\hline
\textbf{Ours\_w/o Atten} &	57.94\%	&72.67\%	&\textbf{77.28\%}	&81.17\%	&82.97\% \\
\textbf{Ours\_NL}	&58.29\%	&73.23\%	&77.42\%	&\textbf{81.37\%}	&82.59\% \\
\textbf{Ours\_SE} &	\textbf{58.92\%}	&\textbf{73.94\%}	&77.22\%	&81.34\%	&\textbf{83.10\%} \\
\hline
\end{tabular}
\label{tab3}
\end{center}
\end{table}

\section{Conclusion}\label{Section 5}
A novel method of obtaining prototypes for metric classification called TrainProto-FSL is proposed in this paper, which contains classifier training and prototypes training. And the whole framework designed as \cite{li2020few} helps our algorithm avoid the shortcomings related with the traditional episodic meta-training.

Experiments show a state-of-the-art performance on a standard vision dataset miniImageNet. It proves that the trained prototypes get from neural network are more efficacious than the hand-designed ones, and the attention mechanism is also useful for the performance improvement. In the following study we will explore the effective self-supervised learning algorithm specially for few shot learning.





\bibliographystyle{IEEEtran}
\bibliography{Few-Shot_Image_Classification_via_Contrastive_Self-Supervised_Learning}
\vspace{12pt}
\end{document}